# Real Time Face Recognition Using Convolutional Neural Networks

Sunil Bhutada
Professor, Dept of IT, SNIST
sunilbhutada@yahoo.com

Rohith Pudari
Dept of IT, SNIST
cnu.sunnypudari@gmail.com

Sai Pavan Mudavath
Dept of IT, SNIST
msaipavan36@gmail.com

*Abstract:*

*Face Recognition is one of the process of identifying people using their face, it has various applications like authentication systems, surveillance systems and law enforcement. Convolutional Neural Networks are proved to be best for facial recognition. Detecting faces using core-ml api and processing the extracted face through a coreML model, which is trained to recognize specific persons. The creation of dataset is done by converting face videos of the persons to be recognized into Hundreds of images of person, which is further used for training and validation of the model to provide accurate real-time results.*

*Keywords: Face Recognition, Convolutional Neural Network, Tensorflow, AlexNet, CoreML Model.*

## Introduction:

Face Recognition is the process of identifying a person using their face. There has been great progress in face recognition due to increase in computation power. As humans have an exceptional ability to recognize faces irrespective of the lighting conditions and varying expressions. The aim of face recognition systems is to surpass the human level of accuracy and speed. This application is divided into two parts:

- Creating a coreml model [1] for recognizing specific people
- Creating an IOS app for taking input from camera and getting outputs from coreml model and displaying the results.

AlexNet [2] is used as the base model for the creation of the core ml model. This model is trained to detect specific faces. . The Ml-kit [3] provided by apple is used to extract faces from the images received from the camera in real-time. The extracted faces are passed through the trained model for prediction and using the ar-kit [4] provided by Apple, the results are displayed on the face using face tracking.

## Literature Survey:

**Convolutional Neural Networks:**
A Convolutional Neural Network (CNN, or ConvNet) are a special kind of multi-layer neural networks, designed to recognize the visual patterns directly from pixel images with minimal pre processing. The models used are trained on ImageNet project [5] which is a large visual database designed for use in visual object recognition software research.

**AlexNet:**
It is a convolutional neural network, designed by Alex Krizhevsky. AlexNet contains eight layers; the first five are convolutional layers, some of them followed by max-pooling layers, and the last three are fully connected layers. It uses the non-saturating Rectified Linear Unit (ReLU) activation function [6], which showed improved training performance over tan-h [7] and sigmoid [8].

**Tensorflow:**
TensorFlow is an open-source software library. TensorFlow is developed by researchers and engineers working on the Google Brain Team for the purposes of conducting machine learning and deep neural networks research, TensorFlow is basically a software library for numerical computation using data flow graphs

**Core-ML:**
Core ML lets you integrate a broad variety of machine learning model types into your app. In addition to supporting extensive deep learning with over 30 layer types, the core-ml models also support standard models such as tree ensembles, Support Vector Machines (SVM), and generalized linear models. Because it's built on top of low level technologies like Metal and Accelerate, Core ML seamlessly takes advantage of the CPU and GPU to provide maximum performance and efficiency. We can run machine learning models on the device so the data doesn't need to leave the device to be analyzed.

## Existing Techniques in face recognition:

**Artificial Neural Networks:**
In this technique Radial basis function artificial neural network is naturally integrated with non-negative matrix factorisation. Also other approaches for process simplification regarding ANNs native linearisation feature and computation speed up.The main disadvantage of this approach is a requirement of greater number of training samples.

**Gabor wavelets [9] :**
The Gabor wavelets exhibit desirable characteristics of capturing salient visual properties like spatial localisation, orientation selectivity and spatial frequency. Different biometrics applications favour this approach.

**D-based face recognition [10] :**
This extends traditional 2D capturing process and has greater potential for accuracy. The depth information does not depend on the pose and illumination making the solution more robust and accurate.

**Video-based recognition [11] :**
This method takes video as input and the main advantage of the approach is the possibility of employing redundancy present in video to improve still image systems. This requires a lot of computational power.

## Architecture:
The proposed architecture of the face recognition in the IOS app which takes its input from the camera i.e., real time images and the input is passed to the vision framework object to check for human faces. The vision framework explores the image and checks for human faces then detects and extracts them in the real time and proceed for prediction of the face. After detecting the faces, these extracted faces are sent the processed input to the coreml model. The core-ml model classifies the image and generates a prediction on the label of the image as per its dataset. The Ar-kit gets the prediction from core-ml model and then displays it to the user in augmented reality by using the face tracking object created by vision framework and displays the label on screen in real time. The following is the architectural diagram of the proposed system.

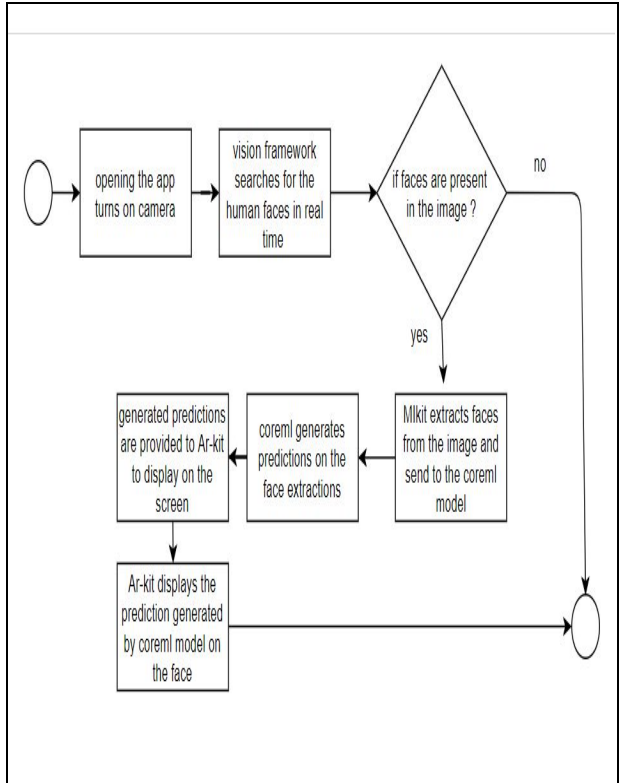

Fig 1. Architecture of the proposed system

**Dataset:**

The dataset comprises of 700 images each under 4 labels of persons to be recognized. To create the training and validation images we made a video containing all the different angles of the face and also with different expressions, then the video is converted to image frame by frame using ffmpeg [12] tool. This is an open source tool takes a snapshot of each frame thus can create about 24 images per sec of a video. Thus, resulting in hundreds of images for a video less than 30 seconds. The images created are uploaded to a cloud storage for easy access, which will be further used for training and validation of the machine-learning model. For each epoch of training, The weights are adjusted accordingly for more efficiency. The distribution of images into training and validation datasets are shown in the following table:

| Label | Training Set | Validation Set |
|---|---|---|
| Rohith | 975 | 325 |
| Pavan | 720 | 180 |
| Vinod | 750 | 250 |
| Indu | 820 | 275 |
| Unknown | 17257 | 1275 |

Table 1. Distribution of the dataset

**Core-ML Model Generation:**

A Alexnet model pre-trained the dataset generated is used to train the model to detect faces. The Alexnet model is trained and optimized for imagenet dataset. It contains eight layers, the first five are convolutional layers, some of them followed by max-pooling layers, and the last three are fully connected layers. It uses the non-saturating Rectified Linear Unit (ReLU) activation function. ReLUs do not require input normalization to prevent them from saturating. However, we still find that the following local normalization scheme aids generalization. is given by the expression β where the sum runs over n "adjacent" kernel maps at the same spatial position, and N is the total number of kernels in the layer. The ordering of the kernel maps is of course arbitrary and determined before training begins. This sort of response normalization implements a form of lateral inhibition inspired by the type found in real neurons, creating competition for big activities amongst neuron outputs computed using different kernels.

Denoting by $a^i_{x,y}$ the activity of a neuron computed by applying kernel i at position (x, y) and then applying the ReLU nonlinearity, the response-normalized activity is $b^i_{x,y}$ is given by the expression:

$$b^i_{x,y} = a^i_{x,y} / \left( k + \alpha \sum_{j=\max(0,i-n/2)}^{\min(N-1,i+n/2)} (a^j_{x,y})^2 \right)^\beta$$

The constants k, n, α, and β are hyper-parameters whose values are determined using a validation set.

**Real-time Faces Extraction:**
The extraction of faces from the camera on a real-time speed is a challenging task. The vision framework [13] created by apple is used to extract faces and also track them in real-time. The face detector object constantly searches for faces. Once a face is detected, then the face is tracked using the track object request.

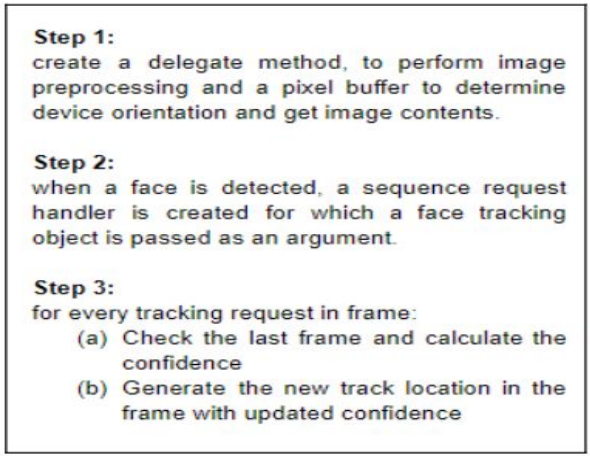

Fig 2. Algorithm to detect and track faces in real-time

**Faces Recognition:**
The faces extracted by the vision framework are passed as an input to the core-ml model which is trained to recognize specific people in an image. The core-ml model recognizes the face and classifies it into a label with its prediction value. Then this generated label is passed to the ar-kit to display the label on the face. The algorithm is as follows

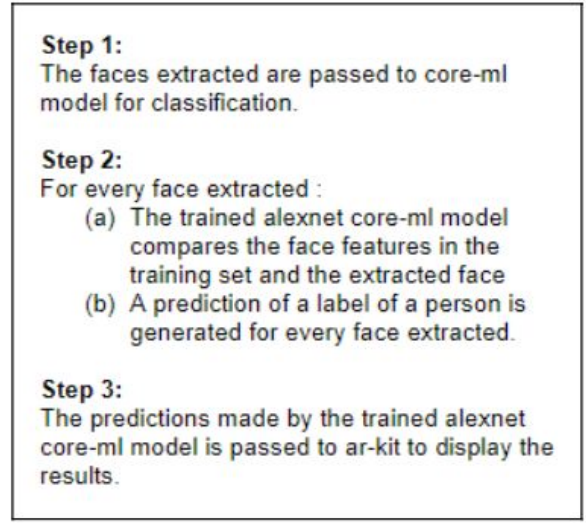

Fig 3. Algorithm to predict labels for faces extracted

**Label Display using AR-Kit:**
Ar-kit is generated by apple to display outputs in Augmented Reality. The results generated by the trained alexnet core-ml model are displayed on the face track object generated by vision framework. A AR session is generated for displaying AR overlay content on the real world screen. We calculate the vector distance between text and face on the screen by using the Pythagorean theorem. When camera captures a human face, it will consider every pixel as a square which are further converted into two right angled triangles. Based on camera angle. The display of the text on the face occurs. The Text will be displayed in augmented reality after the predictions generated by the core-ml model.

**Step 1:**
The labels predicted by core-ml model and the face track object generated by vision framework is passed to Ar-kit

**Step 2:**
The Pythagorean Theorem[] is used to get the angle of the camera on the face and display the text.

**Step 3:**
The text is displayed according to the angle of camera and the face tracking object on the overlaying AR layer.

Fig 4. Algorithm for displaying predictions on the screen.

## Results:

The training of the model is done and the performance metrics of the model on the validation is shown in the following graphical representation.

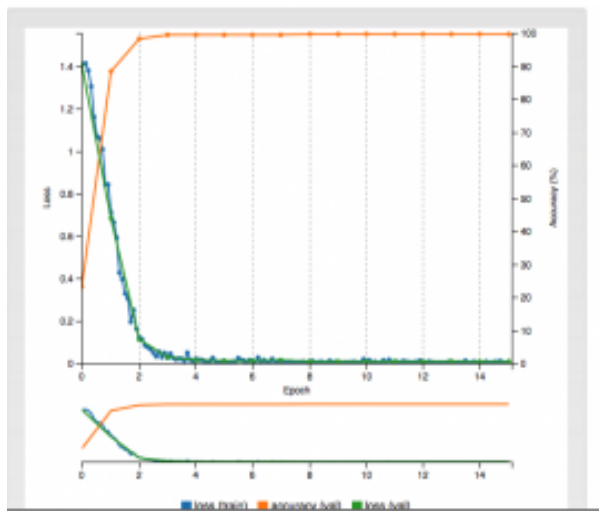

Fig 5. Model evaluation graph

When the model is tested the following results were obtained and we made a test on each epoch to ensure there is no overfitting or underfitting is done. The following table consists of the mean accuracies of all the results obtained for 30 epochs of training.

| Label | Accuracy (on validation data) | Accuracy (on test data) |
|---|---|---|
| Rohith | 92.35% | 94.32% |
| Pavan | 82.94% | 91.74% |
| Vinod | 77.90% | 71.11% |
| Indu | 71.15% | 73.23% |
| Unknown | 84.32% | 87.92% |

Table 2. Mean accuracy results

The total accuracy with all the labels combined was 82.74%. The following image shows a sample prediction on how a image is predicted using the model.

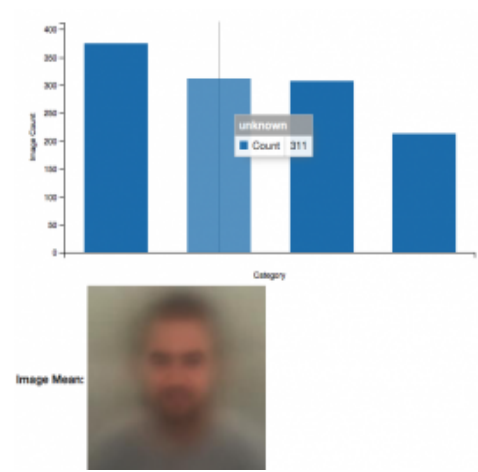

Fig 6. Sample prediction

## Conclusion & Future Scope:

Face recognition has come a long way through this past few years, they are being widely used for security and authentication purposes. As the cost of the equipment is being decreased and more accurate models are being generated, it's been easy to implement the face recognition technology with cost effectiveness and high accuracy results. Image-based face recognition is still a very challenging topic after decades of exploration. A number of typical algorithms are presented, being categorized into appearance-based and model-based schemes while each one of them has their own pros and cons. The proposed algorithm was utilized for face recognition across illumination changes. Although a number of efforts have been made on pose-invariant face recognition, the performance of current face recognition system are still not satisfactory and to be improved on a large margin to reach human level of accuracy.

## References:

[1] Coreml-framework <https://developer.apple.com/documentation/coreml>

[2] Alex Krizhevsky Ilya Sutskever Geoffrey E. Hinton, "ImageNet Classification with Deep Convolutional Neural Networks" journal of neural information processing system, paper: 4824, 2012

[3] Google Ml-kit <https://developers.google.com/ml-kit/>

[4] AR-kit-Framework <https://developer.apple.com/arkit/>

[5] Jia Deng, Wei Dong, Richard Socher, Li-Jia Li, Kai Li and Li Fei-Fei, "ImageNet: A large-scale hierarchical image database", Print ISSN: 1063-6919, IEEE Conference 2009.

[6] Abien Fred M. Agarap, "Deep Learning using Rectified Linear Units (ReLU)", Arxiv:1803.08375v2 [cs.NE] 7 Feb 2011.

[7] Bing Xu, Ruitong Huang, Mu Li, "Revise Saturated-Activation-Functions", arXiv:1602.05980v2 [cs.LG] 2 May 2016.

[8] Masayuki Tanaka, "Weighted Sigmoid Gate Unit for an Activation Function of Deep Neural Network", arXiv:1810.01829v1, 2014

[9] Naghdy, G., Wang, J. & Ogunbona, P. Texture analysis using Gabor wavelets. Proceedings of SPIE - Human Vision and Electronic Imaging (pp. 74-85). The International Society for Optical Engineering. 1996.

[10] Faizan Ahmad, Aaima Najam and Zeeshan Ahmed, " Image-based Face Detection and Recognition", Arxiv Publications, 2013.

[11] O. Arandjelovic, G. Shakhnarovich, J. Fisher, R. Cipolla, and T. Darrell. Face recognition with image sets using manifold density divergence. In CVPR, 2005.

[12] Kyu Yeon Jeon, Jinhong Yang, Sooji Jeon, Sungkwan Jung, "Video conversion scheme with animated image to enhance user experiences on mobile environments", Consumer Electronics 2016 IEEE 5th Global Conference on, pp. 1-2, 2016.

[13] Vision Framework <https://developer.apple.com/documentation/vision>